\pdfoutput=1

\documentclass[11pt]{article}

\usepackage[preprint]{acl}

\usepackage{times}
\usepackage{latexsym}

\usepackage[T1]{fontenc}

\usepackage[utf8]{inputenc}
\usepackage{microtype}

\usepackage{inconsolata}

\usepackage{graphicx}
\usepackage{amsmath}
\usepackage{booktabs}
\usepackage{multirow}
\usepackage{xcolor}
\usepackage{lipsum}
\usepackage{float}
\usepackage{hyperref}
\usepackage{framed}       %
\usepackage{xcolor}       %
\usepackage{mdframed}  
\usepackage{amssymb}

\definecolor{shadecolor}{gray}{0.9}

\title{LLMs Reading the Rhythms of Daily Life: Aligned Understanding for Behavior Prediction and Generation}

\author{
  Fanjin Meng\thanks{\ \ Fanjin Meng and Jingtao Ding contributed equally to this work.} \and 
  Jingtao Ding\thanks{\ \ Corresponding authors.} \and 
  Nian Li \and 
  Yizhou Sun \and 
  Yong Li\footnotemark[2] \\
  Department of Electronic Engineering, BNRist, Tsinghua University, Beijing, China \\
  \texttt{mengfj23@tsinghua.edu.cn} \\
  \texttt{dingjt15@tsinghua.org.cn}, \texttt{liyong07@tsinghua.edu.cn}
}

\begin{document}
\maketitle
\begin{abstract}
Human daily behavior unfolds as complex sequences shaped by intentions, preferences, and context. Effectively modeling these behaviors is crucial for intelligent systems such as personal assistants and recommendation engines. While recent advances in deep learning and behavior pre-training have improved behavior prediction, key challenges remain—particularly in handling long-tail behaviors, enhancing interpretability, and supporting multiple tasks within a unified framework. Large language models (LLMs) offer a promising direction due to their semantic richness, strong interpretability, and generative capabilities. However, the structural and modal differences between behavioral data and natural language limit the direct applicability of LLMs.
To address this gap, we propose Behavior Understanding Alignment (BUA), a novel framework that integrates LLMs into human behavior modeling through a structured curriculum learning process. 
BUA employs sequence embeddings from pretrained behavior models as alignment anchors and guides the LLM through a three-stage curriculum, while a multi-round dialogue setting introduces prediction and generation capabilities.
Experiments on two real-world datasets demonstrate that BUA significantly outperforms existing methods in both tasks, highlighting its effectiveness and flexibility in applying LLMs to complex human behavior modeling. The code is available at \color{blue}\url{https://anonymous.4open.science/r/dasjijio-21B2/}
\end{abstract}

\section{Introduction}

Human daily life unfolds as a sequence of behaviors—ranging from habitual routines to spontaneous actions—each reflecting underlying intentions, preferences, and contextual factors.
Accurately modeling and understanding these human daily behaviors is fundamental to a wide range of intelligent systems, including personalized assistants, recommender engines, and context-aware services~\citep{chung2018intelligent,tulshan2019survey,savcisens2023using}.
Traditional approaches~\citep{zhu2017next,chen2018predictive,yuan2023learning}, particularly those based on deep learning, have primarily focused on behavior prediction: learning to predict the next behavior event based on historical sequences~\citep{kang2018self,sun2019bert4rec}.
Recently, with the increasing availability of large-scale behavioral datasets and inspired by the success of pre-training paradigms in natural language processing (NLP)~\citep{radford2019language,dubey2024llama}, behavior pre-training has emerged as a promising technique.
These methods~\citep{gong2024population,savcisens2024using,zhai2024actions} exploit vast human daily behavioral corpora to capture intricate temporal dependencies and latent patterns, leading to significant improvements in predictive accuracy.

Despite these advances, existing human daily behavior modeling approaches suffer from several \textbf{critical limitations}.
First, they struggle to model \textbf{long-tail behaviors}—actions that occur infrequently or are newly emerging—due to inherent data sparsity issues~\citep{hu2025alphafuse,kimlarge}.
Second, their ``black-box'' nature offers limited insight into the decision-making process, 
creating a gap between the model's predictions and \textbf{human-interpretable reasoning}~\citep{lei2024recexplainer}.
Third, most models are designed for a single task, focusing on either prediction or generation, and lack the flexibility to handle both within a \textbf{unified framework}.

Recent developments in Large Language Models (LLMs) offer a powerful new direction for addressing these challenges.
LLMs provide several distinct advantages:
(1) Their rich semantic representations, learned from vast textual corpora, can enhance the modeling of long-tail behaviors by providing crucial contextual understanding~\citep{liu2024llm,sheng2024language}.
(2) Trained on extensive human-generated text, LLMs can process and articulate behavioral patterns in a textual format that aligns more closely with human cognition, thereby enhancing model interpretability.
(3) Their \textbf{inherent generative capabilities} support multitask learning through natural language, enabling both behavior prediction and generation within a single, unified model.
Overall, integrating LLMs presents a clear opportunity to overcome the core limitations of traditional behavior modeling.

However, human daily behavioral data differs fundamentally from textual data in both structure and modality. As LLMs are trained primarily on natural language, they cannot directly interpret the features or outputs of conventional human daily behavior modeling pipelines. To address this challenge, we propose the Behavior Understanding Alignment (BUA) framework, a novel approach that integrates LLMs into human daily behavior modeling, which first leverages sequence embeddings from a pretrained behavior model as alignment anchors and guides the LLM through a structured three-stage curriculum, designed to progressively bridge the modality gap.
And we further introduce a multi-round dialogue setting that incrementally incorporates prediction and generation, forming a natural reasoning chain: understanding, then predicting and generating, which effectively leveraging the model’s comprehension of human daily behavior sequences to enhance both prediction and generation performance.
The contributions of this work are summarized as follows:

\begin{itemize}
    \item We are the first to propose training an LLM to explicitly \textit{understand} human daily behavior sequences—by aligning behavioral and language modalities—as a foundational step for improving downstream prediction and generation tasks.
    
    \item We introduce the Behavioral Understanding Alignment (BUA) framework, which uniquely combines a three-stage curriculum learning pipeline with a multi-round dialogue mechanism to synergistically enhance the model's capabilities in understanding, predicting, and generating human behaviors.
    
    \item Experimental results on two real-world datasets demonstrate that BUA achieves state-of-the-art performance in both prediction and generation tasks. Comprehensive ablation studies further validate the critical role of our structured curriculum and dialogue-based reasoning process in achieving these results.
\end{itemize}

\section{Related Work}

\subsection{Behavior Modeling}

Modeling daily human behavior hinges on capturing underlying regularities in user behavior sequences, typically through two tasks: behavior prediction and behavior generation. 
Early behavior prediction models, such as TRNN~\citep{zhu2017next}, utilized time-difference-aware embeddings to enhance temporal modeling. 
As datasets expanded, transformer-based pretraining methods like BehaveGPT~\citep{gong2025behavegpt} and Life2Vec~\citep{savcisens2024using} became prevalent, significantly improving predictive accuracy. 
However, these methods often struggle with long-tail behaviors due to limited sample diversity. 
For behavior generation, early rule-based and agent-based models~\citep{kim2019simulating,pfosertowards} relied on hand-crafted logic, limiting their ability to capture real-world complexity. 
SAND~\citep{yuan2023learning} advanced this by using neural stochastic differential equations, enabling more realistic dynamics without fixed rules, though its static generation parameters limit adaptability. 
More recently, D2A~\citep{wang2024simulating} trained an LLM as a cognitively inspired agent guided by a dynamic value system, enhancing behavioral diversity and flexibility. 
However, it underutilizes the LLM's potential for sequence generation based on a deep, multimodal understanding of behavioral context. 

\begin{figure*}[t]
\vspace{-25mm}
    \centering
    \includegraphics[width=0.98\linewidth]{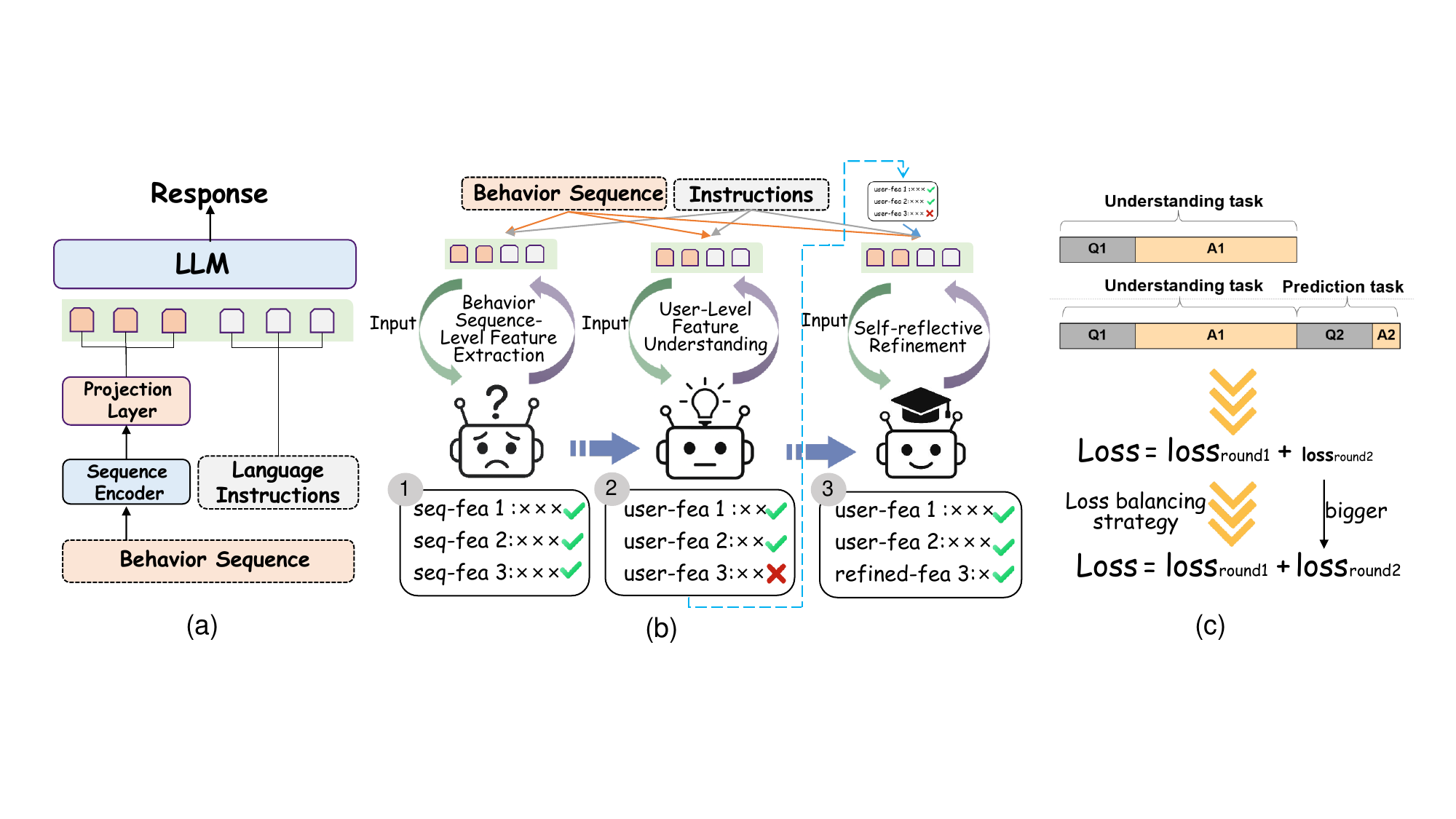}
    \vspace{-22mm}
    \caption{The framework of Behavior Understanding Alignment (BUA). 
    (a) the modality conversion process using sequence embedding. 
    (b) curriculum for Behavior Understanding Alignment: seq-fea, user-fea, and refined-fea represent features learned in Stage 1 (Sequence-Level), Stage 2 (User-Level), and Stage 3 (Self-Reflection). The \textbf{$\checkmark$} and \textbf{$\times$} marks indicate the correctness of the learned features.(c) Understanding-enhanced prediction and generation via multi-round dialogue.}
    \label{fig:framework}
    \vspace{-18pt} 
    \end{figure*} 

\subsection{Alignment in Recommendation} 

Our behavior prediction task is defined as predicting the next behavior based on the user's most recent $L$ behavioral events, which is similar to sequential recommendation. 
The two differ in focus: behavior prediction emphasizes recurring daily behaviors, whereas sequential recommendation often targets novel items. 
To the best of our knowledge, no prior work has focused on explicitly aligning LLMs with the underlying semantic representations of entire behavior sequences, so we draw upon related research in sequential recommendation.
Existing work that incorporates large language models (LLMs) into sequential recommendation can be broadly categorized into two directions: (1) LLMs as standalone recommendation systems~\citep{tan2024idgenrec,kimlarge,liao2024llara,zhang2025collm}, and (2) LLMs as enhancers of traditional systems~\citep{ren2024representation,liu2024llm,hu2025alphafuse,wang2024simulating}.
However, works in the first category do not explore the cross-modal connection or establish a model's explicit understanding of user behavior sequences. Their item-embedding alignment-based approach limits the ability to fully leverage collaborative knowledge.
The latter line of work offers superior inference efficiency compared to LLM-based systems. However, they inherit the limitations of traditional deep learning models, including poor generalization to new items and tasks, and limited interpretability due to their black-box nature\cite{lei2024recexplainer}.

\section{Methodology}

\subsection{Problem Formulation}
Let $x_i$ denote a basic behavioral event, represented as a four-tuple $(d_i, t_i, l_i, b_i)$, where $d_i$ is the Day of week index, $t_i$ the Timestamp ID, $l_i$ the Location ID, and $b_i$ the Behavior Type ID. The behavior type captures high-level daily activities—such as exercising or gaming—rather than fine-grained actions.
We consider two related tasks under this representation:

\textbf{Behavior Prediction and Generation.}
Given a user’s recent behavior sequence $X_{\text{seq}} = \{ x_1, x_2, \dots, x_{L_1} \}$, the model is tasked with either (1) predicting the next behavior $b \in \mathcal{B}$, or (2) generating a future sequence $Y_{\text{seq}} = \{ y_1, y_2, \dots, y_{L_2} \}$, where each $y_i$ is a four-tuple $(d_i, t_i, l_i, b_i)$.

\subsection{Overview}
\label{sec:overview}

The framework of our method, Behavior Understanding Alignment (BUA), is illustrated in Figure~\ref{fig:framework}. 
To effectively bridge the gap between numerical behavior data and textual semantic space, we first introduce a \textbf{Sequence Embedding Alignment} module, as detailed in Figure~\ref{fig:framework}(a).
Given a user behavior sequence $X_{\text{seq}}$, we utilize behaveGPT~\citep{gong2025behavegpt}, a model pretrained on large-scale behavioral data, as the behavior encoder $g_\phi$. The penultimate hidden state is projected into a unified representation space via a lightweight two-layer MLP, producing the behavior sequence embedding $H_{\text{seq}}$:
\begin{equation}
    H_{\text{seq}} = \text{MLP}(g_\phi(X_{\text{seq}}))
\end{equation}
This embedding $H_{\text{seq}}$ serves as a \textbf{cross-modal anchor}, enabling the LLM to perceive and reason about continuous behavior signals alongside textual instructions.

Building upon this alignment foundation, we propose a three-stage structured curriculum to progressively enhance behavior understanding: (1) Sequence-Level Understanding, (2) User-Level Feature Modeling, and (3) Self-Reflective Refinement. 
In the second stage, we additionally incorporate a multi-round dialogue setup that integrates prediction and generation tasks.
Throughout these stages, the model is optimized by maximizing the likelihood of the target output $y$ given the multimodal input (text instruction $X_{\text{Ins}}$ and behavior embedding $H_{\text{seq}}$):
\begin{equation}
\mathcal{L} = -\frac{1}{N} \sum_{i=1}^{N} \log P_\theta\left( y_t \mid y_{<t}, X_{\text{Ins}}, H_{\text{seq}} \right),
\end{equation}
where $N$ denotes the length of the response $y$.

\subsection{Curriculum for Behavior Understanding Alignment}
We propose a three-stage curriculum for user behavior understanding tasks, progressing from simple to complex. The first stage targets sequence-level understanding, the second emphasizes user-level feature modeling, and the third incorporates self-reflective refinement to further enhance behavioral representations. All tasks are detailed in Appendix~\ref{app:alltask}.

\subsubsection{Stage 1: Behavior Sequence-Level Feature Extraction}

In this initial stage, the model learns to interpret behavior sequence embeddings from language modalities, building a foundation for deeper understanding of user daily behaviors in later stages. Based on empirical insights, we introduce three basic tasks:
\begin{itemize}
    \item \textbf{Historical Sequence Reconstruction}: The model reconstructs the original behavioral sequence in natural language, capturing key temporal, spatial, and behavioral transitions. This event-level task establishes the groundwork for multimodal understanding.

    \item \textbf{Current Scene Summary}: The model summarizes the user’s recent context over the past two hours (e.g., "morning commute"), requiring it to extract and generalize key patterns, advancing its sequence-level understanding.
  
    \item \textbf{Future Scene Inference}: The model predicts the user’s likely context in the next two hours (e.g., "evening commute" or "pre-bedtime leisure"), demonstrating its ability to analyze sequence dynamics and temporal trends.
\end{itemize}
In practice, we also introduce simple user-level inference tasks at this stage, such as home/workplace location identification and user hobby inference, which provide a natural transition to the more complex user-level understanding required in Stage 2. 
To enhance training efficiency, we freeze both the sequence encoder and the LLM during this stage, allowing only the parameters of the projection layer (MLP) to remain trainable, following the common acceleration training techniques for large multimodal language models~\cite{liu2023llava}.

\subsubsection{Stage 2: User-Level Feature Understanding} 
In this phase, the focus shifts to capturing deeper user-level features embedded within behavior sequences. It is not sufficient for the model to recognize superficial transitions between time, location, or activity; instead, it must extract the user features that characterize the underlying mechanisms behind observed behaviors. Such features are critical for understanding behavioral regularities and informing prediction and generation tasks.
Inspired by how humans infer features regarding users’ lifestyle in terms of what they behave, we design the following tasks to guide user-level feature learning:
\begin{itemize}
    \item \textbf{User Key Behavior Identification}:  The model identifies semantically rich behaviors that are frequent or mark transitions between daily phases (e.g., taking the subway after work indicates a shift from work to evening leisure). These behaviors are critical for inferring user intent evolution.
    
    \item \textbf{User Behavior Pattern Discovery}:  The model detects recurring behavioral subsequences and consistent temporal-spatial patterns (e.g., watching TikTok during commutes). These patterns reveal deeper user preferences and routines.
    
    \item \textbf{User Feature Summarization}:  The model abstracts high-level user features (e.g., This user prefers light entertainment during his evening commute), which provides a higher-level, more essential understanding of user behavior features.
\end{itemize}
These tasks are intentionally sequenced from simple to complex, forming a structured learning path. To support effective user-level understanding, this hierarchical curriculum is enforced during training: the model must perform User Key Behavior Identification first, followed by Behavior Pattern Discovery, and finally User Feature Summarization. 
Additionally, during training at this stage, we freeze the sequence model parameters while allowing the parameters of the projection layer and LLM to be adjustable.

\subsubsection{Stage 3: Self-reflective Refinement}
In this phase, we introduce Self-Reflective Refinement to enhance the model’s understanding of user features. Despite the training in the first two phases, the performance on the User Feature Summarization task remained suboptimal. We evaluated the generated summaries and manually inspected low-scoring responses, finding that there are some recurring issues, such as unclear relationships between behavioral features.
As these issues did not stem from a fundamental misunderstanding, the model might actually have developed a comprehensive understanding of user features during the Sequence-Level Understanding and User-Level Feature Extraction stages. Instead, the model only requires more "thinking" to generate reasonable and accurate user features.
To address this, we propose a self-reflective iteration strategy that empowers the model to identify and correct its own shortcomings. Specifically, we summarize the recurring issues, and design targeted correction criteria, guiding the model to review and revise its earlier outputs based on clear feedback. The task is defined as follows:
\begin{itemize}
    \item \textbf{Self-Reflective Refinement}: The model reviews low-quality user feature summaries, identifies key issues, and refines them using its understanding of sequence embeddings, producing more accurate and coherent summaries.
\end{itemize}
Based on the foundations established in the first two stages, this strategy leverages the model's reasoning capabilities for iterative improvement, resulting in more robust and accurate user feature representations. During this phase, we freeze the sequence encoder and projection layer while keeping LLM parameters trainable.

\subsection{Multi-round Dialogue Setup}
To develop behavior prediction and generation capabilities, we introduce a multi-round dialogue framework in the second stage of User-Level Feature Understanding. This approach enables the model to simultaneously refine prediction and generation skills while deepening its understanding of user behavior features.
In this setup, the model starts with the Key Semantic Behavior Recognition task in the first round, progressively completes all user-level behavior understanding tasks, and concludes with the corresponding prediction or generation tasks in the final round. By leveraging intermediate understanding and analysis from earlier rounds, the model enhances the accuracy and effectiveness of downstream tasks.
The optimization loss for this multi-round dialogue setting is defined as: 

\begin{equation}
\begin{split}
    Loss
    = \mathrm{\frac{ 1 }{\sum_{i=1}^N T_i}}\bigg( & \sum_{i=1}^N \sum_{t=1}^{T_i} \log P_\theta\big( y^{(i)}_t \big| \\
    & y^{(i)}_{<t}, X^{(i)}_{\text{Ins}}, Y^{(i)}, X_{\text{seq}} \big) \bigg)
    \label{eq:multi-turn-loss},
\end{split}
\end{equation}
where $N$ is the total number of dialogue rounds, $T_i$ is the number of tokens in the answer for the $i$-th round, $\theta$ denotes the LLM parameters,$y^{(i)}_t$ is the $t$-th token of the $i$-th round’s output, $X^{(i)}_{\text{Ins}}$ is the input Instruction, and $Y^{(i)}$ represents the corresponding answer for the $i$-th round.

However, multi-round dialogues risk imbalanced training across different rounds. Rewriting the loss from the token level to the round level, we get
\begin{equation} %
    Loss_{\text{multi-turn}} 
    = -\sum_{i=1}^N\frac{ T_i }{\sum_{i=1}^N T_i} loss_{\text{i}},
    \label{eq:rewriting loss} %
\end{equation}
where $loss_i$ denotes the average loss for the $i$-th round. This means that rounds with longer answers dominate the overall loss, while those with shorter outputs receive less attention at the round level. This is problematic in practice, as understanding tasks typically involve long outputs (often exceeding 100 tokens), whereas prediction tasks only output the predicted behavior type (often fewer than 5 tokens). As a result, the model struggles to effectively learn shorter prediction tasks.

To address this issue, we introduce a simple yet effective loss balancing strategy that ensures equal attention across rounds. 
Specifically, we apply a weight $W_i=\frac{\sum_{i=1}^N T_i}{N T_i}$ that is inversely proportional to the length of the answer in each round, encouraging balanced learning across all rounds. The final loss function becomes:
\begin{equation}
\begin{split}
    Loss_{\text{weighted}} 
    &= -\frac{1}{N} \sum_{i=1}^N 
       \log P_\theta \Bigl( y^{(i)}_t \Bigm| \\
    &\quad y^{(i)}_{<t},\ X^{(i)}_{\text{Ins}},\ Y^{(i)},\ X_{\text{seq}} \Bigr)
\end{split}
\end{equation}
This balanced loss formula significantly enhances the performance of the model on the behavior prediction task without significantly reducing the effectiveness on the understanding task.

\section{Experiments}
\subsection{Experimental Settings}

\textbf{Datasets}. 
We evaluated our model on two real-world user behavior datasets:
\textbf{Behavior dataset}: 
This dataset is derived from the user's mobile phone logs. After desensitization, it includes 37 daily behaviors that cover a wide range of life scenarios, including activities related to learning, work, leisure, and more.
\textbf{Tencent Dataset~\citep{shao2024beyond}}:
This dataset is derived from the user's social network and the user's movement trajectory. It includes 14 human behavior intentions, such as eating, going home, working, etc.

For both datasets, we split the users in a ratio of 8:1:1 to create training, validation, and test datasets. For more detailed information about the datasets and their splits, please refer to the Appendix~\ref{app:dataset}.

\begin{table*}[t]
\centering
\vspace{-3mm}
\caption{Experiment Results on Next Behavior Prediction}
\vspace{-3pt} 
\makebox[\textwidth]{
    \scalebox{0.75}{
    \setlength{\tabcolsep}{4pt}
    \renewcommand{\arraystretch}{1.3}
\begin{tabular}{c|c| c c c c c c | c c c c c c} 
 \toprule
 \midrule
 \multirow{2}{*}{\textbf{Category}} &\multirow{2}{*}{\textbf{Method}} & \multicolumn{6}{c|}{\textbf{Behavior Dataset}} & \multicolumn{6}{c}{\textbf{Tecent Dataset}} \\
\cmidrule(lr){3-8} \cmidrule(lr){9-14}
 & & $Rec_w$ & $Prec_w$ & $Overall$ & $Head$ & $Medium$ & $Tail$ & $Rec_w$ &  $Prec_w$& $Overall$ & $Head$ & $Medium$ & $Tail$\\

 \midrule
\multirow{2}{*}{Traditional} 

& SASRec & 0.546 & 0.535 & 0.291 & 0.420 & 0.340 & 0.222  
 &0.328 & 0.269 &0.097& 0.29 & 0.045 & 0.021 \\
&behaveGPT  & 0.567 & 0.551 & 0.206 & 0.442 & 0.354 & 0.027  & 0.509 & 0.426 & 0.113 & 0.537 & 0.000 & 0.000\\
\midrule

\multirow{3}{*}{LLM-Enhanced} 
& PITuning & 0.617 & 0.603 & \underline{0.408}& \underline{0.481}& 0.444 & 0.361& 0.524& 0.466 & 0.120& 0.546 & 0.009 &0.000

 \\
& AlphaFuse  & 0.578 & 0.575 & 0.242 & 0.457 & 0.380 & 0.075 & 0.507 & 0.435 & 0.118 & \underline{0.547} & 0.001 & 0.000 \\
\midrule

\multirow{6}{*}{LLM-Based} 
& Deepseek-V3 & 0.492 & 0.495 & 0.237 & 0.330 & 0.265 & 0.191 & 0.318 & 0.282 & 0.119 & 0.303 & \underline{0.083} & \underline{0.038} \\
& TALLRec & 0.617 & 0.607 & 0.398& 0.452 & 0.434 & 0.355  &0.561 & 0.543 &0.134& 0.513 & 0.044 & 0.019
 \\
& A-LLMRec & 0.584 & 0.557 & 0.348 & 0.422 & 0.394 & 0.299 & 0.539 & 0.523 & 0.140 & 0.542 & 0.037 & 0.025 \\  %
& CoLLM & \underline{0.618}& 0.596 & \underline{0.408}& 0.453 & \underline{0.448} & \underline{0.363} & 0.560 & 0.543 & \underline{0.152}& 0.530 & 0.067 & 0.034
 \\ 
& LLaRA & 0.615 & \underline{0.608}& 0.404 & 0.462 & 0.439 & 0.361& \underline{0.564} & \underline{0.545} & \underline{0.152}& 0.543 & 0.060 & 0.033
 \\
& \textbf{BUA} &\textbf{0.644}&\textbf{0.642}&\textbf{0.471}&\textbf{ 0.538 }&\textbf{0.489}&\textbf{0.446}&\textbf{0.600}& \textbf{0.574} & \textbf{0.207}& \textbf{0.62} & \textbf{0.114} &\textbf{0.041}
 \\
\midrule
&Improv &  4.2\%& 5.6\%& 15.4\%& 11.9\%& 9.2\%& 22.9\%& 6.4\%& 5.3\%& 36.2\%& 13.4\%& 37.4\%&7.9\%\\
\bottomrule
\end{tabular}}
    \vspace{-5mm}
    \label{tab:overall}}
\end{table*}

\begin{table*}[t]
\centering
\vspace{-3pt} 
\caption{Experiment Result on Behavior Sequence Generation}
\vspace{-3pt} 
\label{tab:metrics}
\makebox[\textwidth]{
        \scalebox{0.85}{
        
        \begin{tabular}{l *{3}{ccc}}
        \toprule
        \multirow{2}{*}{Method}  & 
        \multicolumn{3}{c}{Event} & 
        \multicolumn{3}{c}{Timestamp} & 
        \multicolumn{3}{c}{Location} \\
        \cmidrule(lr){2-4} \cmidrule(lr){5-7} \cmidrule(lr){8-10}
         & $\text{Bleu} \uparrow$ & $\text{TVD} \downarrow$ & $\text{JSD} \downarrow$ & $\text{Bleu} \uparrow$ & $\text{TVD}\downarrow$ & $\text{JSD} \downarrow$ & $\text{Bleu} \uparrow$ & $\text{TVD} \downarrow$ & $\text{JSD} \downarrow$ \\
        \midrule
        behaveGPT & 0.009 & 0.945 & 0.632 & — & — & — & — & — & — \\
        SAND         & 0.142 & 0.304 & 0.083 & 0.344 & 0.204 & 0.038 & — & — & — \\
        D2A   & 0.315 & 0.183 & 0.039 & 0.287 & 0.223 & 0.049 & 0.396 & 0.529 & 0.173 \\
        Ours         & 0.354 & 0.140 & 0.020 & 0.541 & 0.147 & 0.020 & 0.711 & 0.065 & 0.005 \\
        \bottomrule
        \end{tabular}
        }
}
\label{tab:overall-gene}
\vspace{-16pt} 
\end{table*}

\textbf{Evaluation Metrics}. For \textbf{behavior prediction task}, we adopt commonly used metrics, weighted precision ($Prec_w$) and weighted recall ($Rec_w$)(equivalent to HR@1), to evaluate the overall prediction performance of the model.
Additionally, user data is often unevenly distributed and exhibits a clear long-tail distribution in practice~\citep{kimlarge}. Following relevant work~\citep{liu2019large,shi2024long}, we construct a long-tail intent test set and adopt global accuracy(denoted as $Overall$), high-frequency behavior accuracy(denoted as $Head$), medium-frequency behavior accuracy(denoted as $Medium$), and long-tail behavior accuracy (denoted as $Tail$) to fairly evaluate the model’s performance across different behavior categories.
For the calculation methods and additional details on all six metrics, please refer to the Appendix~\ref{app:metric-pred}.
For \textbf{behavior generation task}, we adopt commonly used metrics $BLEU$, $TVD$, and $JSD$ to measure the time, location, and behavior similarity between the generated sequence and the real sequence data. The calculation methods for the metrics are outlined in the Appendix~\ref{app:metric-gene}.

\textbf{Baselines}.
For \textbf{behavior prediction task}, We selected representative algorithms from various categories to compare with our proposed algorithm. For traditional deep learning methods, we chose SASRec~\citep{kang2018self}, BehaveGPT~\citep{gong2025behavegpt}. For pure LLM-based prediction methods, we selected DeepSeek-V3~\citep{deepseekai2025deepseekv3technicalreport} and TallRec~\citep{bao2023tallrec}. For methods that use modality fusion and LLM as recommendation systems (similar to ours), we selected A-LLMRec~\citep{kimlarge}, CoLLM~\citep{zhang2025collm}, and LLaRa~\citep{liao2024llara}. For methods that employ modality fusion and LLM as recommendation system enhancers, we selected PI-Tuing~\citep{gong2024population} and AlphaFuse~\citep{hu2025alphafuse}.
For \textbf{behavior generation task}, We chose SAND~\citep{yuan2023learning}, a representative method based on deep learning, and  D2A~\citep{wang2024simulating}, which uses LLM for user behavior activity generation based on Maslow's Theory.
For further details on the baselines, please refer to the Appendix~\ref{app:baselines}.

\textbf{Implementation Details}. 
The hardware used in this experiment consists of 8 NVIDIA A100 40G GPUs. We chose Qwen2.5-7B~\citep{qwen2025qwen25technicalreport} as the backbone model for the experiment.
More details are in the Appendix~\ref{app:Implementation}.

\subsection{Overall Performance}
\subsubsection{Behavior Prediction Experiment}
We evaluated our model on two real-world datasets against all baselines. Our model consistently outperformed all baselines across all metrics, confirming its effectiveness, as shown in Figure~\ref{tab:overall}. Further insights are detailed below.

\textbf{Overall Comparison}. 
The results show that our method(BUA) outperforms all baselines on both $Prec_w$ and $Rec_w$ under the real data distribution. Additionally, most LLM-based methods surpass traditional models like SASRec, underscoring the value of semantic information in behavior prediction. Notably, item embedding fusion approaches (e.g., LLaRA, A-LLMRec) offer no clear advantage over the pure LLM method, TallRec, indicating that item embeddings alone are insufficient to fully leverage sequential knowledge.

\textbf{Different Categories of Behaviors Comparison}. 
The results show that our method(BUA) achieves substantial gains across high-frequency, medium-frequency, and long-tail behaviors, with an overall average improvement of 25.8\% over the best baseline on both datasets. While the LLM-based TallRec outperforms SASRec by over 50\% on long-tail behaviors, it shows only a 10\% gain on high-frequency ones on Behavior dataset, emphasizing the importance of semantic information for long-tail prediction. Although the item-embedding fusion method LLaRA slightly outperforms the pure LLM TallRec, the margin is small compared to BUA, further confirming the advantage of BUA.

\subsubsection{Behavior Generation Experiment}

We evaluated our method on the Behavior dataset against all baselines. To ensure fair comparison with SAND, which outputs fixed-time behaviors, we generated one day of future behaviors. Our model consistently outperformed all baselines across all metrics, demonstrating strong robustness and effectiveness (see Figure~\ref{tab:overall-gene}). Note that $-$ in table indicates the model lacks generation capability and is not applicable for evaluation.
Additionally, we found that BehaveGPT, despite large-scale pretraining, performs poorly on the generation task even after fine-tuning, revealing limited capability. While SAND generates more accurate timestamps than D2A, it lags in behavior accuracy. These results underscore the importance of behavioral semantics and the difficulty LLMs face with temporal and numerical features. Our method addresses this by first understanding and summarizing behavioral patterns, leading to superior performance. 
In addition, we conducted experiments using the generated data on downstream prediction tasks, refer to Appendix~\ref{app:Practical Applications} for details.

\begin{table}[t]
\caption{Cross-Model Enhancement via Behavioral Understanding Transfer}
\vspace{-5pt} 
\label{tab:cross}
\centering  %
\setlength{\tabcolsep}{4pt}  %
\renewcommand{\arraystretch}{1.1}  %
\resizebox{0.48\textwidth}{!}{  %
\begin{tabular}{c|cc|cccc}  %
\toprule
\textbf{Method} & $Rec_w$ & $Prec_w$ & $Overall$ & $Head$ & $Medium$ & $Tail$ \\
\midrule
Tallrec         & 0.609 & 0.587 & 0.384 & 0.460 & 0.440 & 0.319 \\
Tallrec-cross   & 0.620 & 0.610 & 0.405 & 0.467 & 0.463 & 0.341 \\
Llara           & 0.605 & 0.585 & 0.362 & 0.460 & 0.412 & 0.296 \\
Llara-cross     & 0.610 & 0.595 & 0.403 & 0.468 & 0.466 & 0.335 \\ 
\bottomrule
\end{tabular}
}
\vspace{-18pt} 
\end{table}

\subsection{Cross-Model Enhancement via Behavioral Understanding Transfer}

To further validate the effectiveness of our model’s behavior understanding, we evaluate whether its extracted user features could enhance other models. These features were added to TallRec and LLARA, resulting in TallRec-cross and LLARA-cross. Specifically, BUA generated user summaries for 20,000 samples to supplement each model’s input. As shown in Table~\ref{tab:cross}, this consistently improved prediction performance, with notable gains on long-tail behaviors.

\begin{table}[h]
    
    \caption{Ablation Study for Behavior Prediction Task}
    \vspace{-3pt} 
    \label{tab:overall-albation_pred}
    \setlength{\tabcolsep}{3pt}
    \resizebox{0.48\textwidth}{!}{
        \begin{tabular}{c|c c|c c c c}
            \toprule
            \textbf{Method} & $Rec_w$ & $Prec_w$ & $Overall$ & $Head$ & $Medium$ & $Tail$ \\
            \midrule
            Ours & 0.644 & 0.642 & 0.471 & 0.538 & 0.489 & 0.446 \\
            w/o stage1 & 0.613 & 0.607 & 0.370 & 0.485 & 0.452 & 0.286 \\
            w/o stage2 & 0.587 & 0.577 & 0.456 & 0.400 & 0.159 & 0.291 \\
            w/o stage3 & 0.592 & 0.586 & 0.334 & 0.420 & 0.357 & 0.300 \\
            w/o loss balance & 0.560 & 0.552 & 0.285 & 0.385 & 0.364 & 0.197 \\
            item-emb & 0.631 & 0.626 & 0.465 & 0.523 & 0.491 & 0.437 \\
            \bottomrule
        \end{tabular}
    }
    \vspace{-18pt} 
\end{table}

\subsection{Ablation Study}
We conduct an ablation study on the Behavior dataset to evaluate the influence of different design components on overall performance. Specifically, we assess the model's performance under the following conditions: (1) Removal of the first stage: Sequence-Level Feature Understanding (w/o stage1), (2) Removal of User-Level Feature Extracting in second stage (w/o stage2), and (3) Removal of the third stage: Self-reflective Refinement(w/o stage3).

\textbf{Behavior Prediction} 
To further analyze the contributions of each component to the Behavior Prediction task, we additionally evaluate:
(4) Removal of loss balancing strategy of multi-turn dialogue in the second stage (w/o loss balance), (5) Use of item embedding instead of sequence embedding for modality alignment (item-emb).
The results are shown in Table~\ref{tab:overall-albation_pred}, with key findings as follows:
All components contribute to overall performance, with the loss balancing strategy in the second-stage multi-turn dialogue having the most significant impact on prediction. Without it, the model favors the understandng task due to its longer output, reducing prediction effectiveness.
Although the first and third stages do not directly target prediction, they improve behavior sequence understanding, indirectly enhancing prediction. Finally, replacing sequence embeddings with item embeddings for modality fusion leads to performance degradation, confirming the superiority of sequence-level representations. Furthermore, we conducted a more detailed analysis of the sources of model errors, which can be found in the appendix~\ref{app:error_source}.

\textbf{Behavior Generation}
All design components contribute to the model’s overall performance. Removing the user-level feature extraction in the second stage has the greatest impact, showing that explicitly generating user behavior features enhances understanding and guides future behavior generation. Additionally, the behavior understanding tasks in the first and third stages improve the model’s grasp of behavior sequences, as their removal weakens pattern understanding and indirectly reduces generation performance. Please refer to Appendix~\ref{app:Ablation-gene} for detailed results.

\subsection{Qualitative and Quantitative Analysis of Interpretability} 
To comprehensively evaluate interpretability, we conducted both qualitative case studies and quantitative human evaluations.

\textbf{Qualitative Analysis: Evolution of User Profiling.} 
We demonstrate the model's ability to capture semantic modalities through the evolution of User Feature Summarization across the three curriculum stages. For clarity, we present only a representative sub-feature, as full outputs are too lengthy.
\begin{center}
\begin{minipage}{0.92\linewidth}
\begin{shaded}
\textit{\textcolor{blue}{\textbf{Stage1:}}The user frequently reads news throughout the day and is a news enthusiast.} \\
\textcolor{blue}{\textbf{Stage2:}}The user has a strong habit of consuming news, \textbf{often checking it multiple times in quick succession}, suggesting \textbf{a desire to stay informed} about current events. \\
\textcolor{blue}{\textbf{Stage3:}}\textbf{Information-Seeking Behavior:} The user has a strong habit of \textbf{consuming news} and \textbf{checking the weather}, indicating \textbf{a desire to stay informed} about current events and environmental conditions. This behavior is \textbf{consistent throughout the week, with slight variations in timing}.
\end{shaded}
\end{minipage}
\end{center}
The outputs demonstrate a clear progression. Stage 1 produces simple summaries (e.g., "likes news"). Stage 2 advances to identifying behavioral patterns (e.g., "checking multiple times in rapid succession") and inferring intent. Finally, Stage 3 generates comprehensive features by abstracting actions into high-level categories, grouping behaviors such as "reading news" and "checking the weather" into higher-level "information-seeking" with richer detail, confirming the effectiveness of our design.

\textbf{Quantitative Analysis: Human Evaluation.}
Following standard protocols~\citep{lei2024recexplainer}, we assessed feature quality and interpretability using 120 random samples from the Honor dataset. We compared features from three sources: (1) Human annotators, (2) BUA (Ours), and (3) the Base Model (Qwen2.5-7B w/o fine-tuning). Blinded samples were scored on a 0–3 scale based on two dimensions:\textbf{Rationality} (The degree to which the features align with the user’s historical behavior) and \textbf{Interpretability} (The extent to which the features help explain the user's future behavior).

\begin{table}[t]
    \centering
    \caption{Average Scores of Human Evaluation}
    \vspace{-3pt} 
    \label{tab:human_eval_avg1}
    \resizebox{0.33\textwidth}{!}{
    \begin{tabular}{lcc}
        \toprule
        Type    & Rationality & Interpretability \\
        \midrule
        Human   & 2.51        & 2.55             \\
        BUA     & 2.46        & 2.39             \\
        No Tune & 1.90        & 1.83             \\
        \bottomrule
    \end{tabular}
    }
    \vspace{6pt}  
    
    \caption{Best Score Probability per Sample}
    \vspace{-3pt} 
    \label{tab:human_eval_best1}
    \resizebox{0.48\textwidth}{!}{
    \begin{tabular}{lccc}
        \toprule
        Category                 & Human & BUA   & No Tune \\
        \midrule
        Best in Rationality      & 39.8\% & 37.9\% & 22.3\%   \\
        Best in Interpretability & 45.3\% & 33.3\% & 21.4\%   \\
        \bottomrule
    \end{tabular}
    }
    \vspace{-15pt} 
\end{table}

As shown in Tables~\ref{tab:human_eval_avg1} and~\ref{tab:human_eval_best1}, BUA significantly improves upon the Base Model (raising interpretability from 1.83 to 2.39), and achieves near-human level performance in "best score probability". These results demonstrate that BUA attains a deep behavior understanding approaching human levels. For the more detailed experimental setup and analysis, please refer to Appendices~\ref{app:human_evaluation} and~\ref{app:Questionnaire_setup}.

\section{Conclusion}

We present Behavior Understanding Alignment (BUA), which leverages structured curriculum learning to align behavior and language modalities via sequence embeddings. This approach effectively unifies prediction and generation within a single framework. Experiments confirm that BUA outperforms existing baselines, achieving state-of-the-art results on two datasets.

\section*{Limitations}

\textbf{Reliance on pre-trained models:}
Our proposed method relies on pretrained behavioral models (e.g., behaveGPT) to produce sequence embeddings, making its performance dependent on the quality and domain coverage of these models, which may limit generalizability in low-resource or cross-domain settings. 

\textbf{Joint task optimization}
While our proposed method BUA introduces prediction and generation tasks through multi-round dialogue, we observed a mismatch in their convergence rates during training: the generation task typically plateaued earlier, while the prediction task continued improving. As a result, selecting a checkpoint based on average loss often leads to suboptimal performance for both tasks. Addressing this convergence imbalance remains an open challenge for future work.

\bibliography{ref}

\clearpage
\appendix

\section{Summary of all understanding tasks}
\label{app:alltask}
\begin{figure}[H]
    \centering
  \includegraphics[width=0.8\columnwidth]{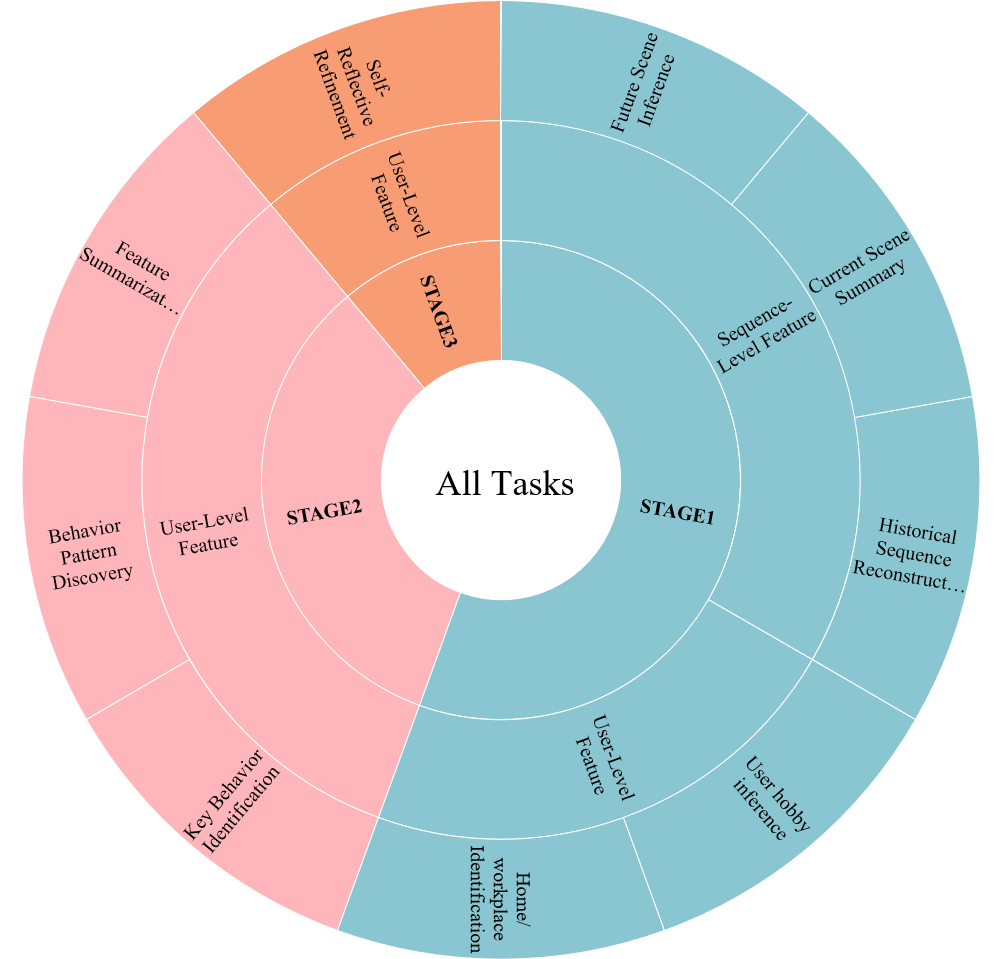}
  \caption{Summary of all understanding tasks}
  \label{fig:all_tasks_huan}
\end{figure}

Summary as shown in Figure~\ref{fig:all_tasks_huan}.

\section{Dataset Information}
\label{app:dataset}
\textbf{Behavior dataset}: This large-scale dataset is derived from mobile phone usage logs. When users interact with their mobile phones, various types of logs are generated, desensitized, and reported with user consent. After desensitizing the original data, we extract 37 daily behaviors that are reliably extracted from raw logs and also cover broad life scenarios, including activities related to learning, work, entertainment, leisure, etc. The dataset spans from March 1, 2024, to April 29, 2024, and consists of over 50 million behavior events from 24,133 anonymous users. We preprocess the dataset and construct samples in the format of $(weekday, timestamp, location, behavior type)$. Since our target is fine-tuning the LLM instead of training from scratch, we only randomly select a subset~(100,000) of data for experiments.

\textbf{Tencent dataset}: The Tencent Trajectory dataset, collected from a social network, captures user mobility trajectories. Points of interest (POIs) are manually annotated with detailed intent types. The dataset includes 2,000 users and spans from October 8 to December 31, 2019, comprising 320,516 records. Each record contains an anonymized location ID, the associated intent type, and a timestamp.

The characteristics of both datasets are presented in Table~\ref{datasets-info}.
\begin{table}[h]
  \centering  %
  \caption{Statistics of the datasets}
  \label{datasets-info}  %
  \setlength{\tabcolsep}{5pt}  %
  \resizebox{0.48\textwidth}{!}{  %
  \begin{tabular}{@{}cccc@{}}  %
    \toprule
    Dataset         & \# Users & \# Behav. Type & \# Logs  \\
    \midrule
    Behavior Dataset & 24,133  & 37            & 100,000 \\
    Tencent Dataset  & 2,000    & 14            & 320,516 \\
    \bottomrule
  \end{tabular}
  }
\end{table}

\textbf{Data Privacy and Anonymization Protocols} %

To ensure the ethical handling of user data and strict adherence to privacy standards, we have implemented the following protection measures:

\begin{itemize}
    \item \textbf{Anonymization Procedures}: We employ a rigorous ID anonymization mechanism. Each user is assigned a randomized anonymous identifier, which is \textbf{periodically refreshed} (e.g., every three months) to prevent long-term user tracking and linkage attacks.
    
    \item \textbf{Location Data Obfuscation}: We strictly avoid collecting precise geospatial coordinates. Instead, we utilize a rule-based aggregation algorithm prior to data upload. This algorithm identifies the user's top-10 frequent activity zones and maps them to \textbf{low-resolution area IDs}. Consequently, only these abstract representations—rather than raw latitude and longitude data—are stored, effectively preserving location privacy.
    
    \item \textbf{Long-Term Privacy Assurance}: Collectively, these measures render it computationally infeasible to re-identify specific individuals through longitudinal behavioral analysis or location inference.
\end{itemize}

\section{Details of Used Metrics for Prediction}
\label{app:metric-pred}

\subsection{Behavior category classification}
For \textit{long-tailed learning}, following the settings of relevant work~\citep{liu2019large,shi2024long}, we evaluate four accuracy metrics based on behavior occurrence frequencies: category-average accuracy across all behaviors ($Overall$), head-category accuracy for behaviors with $>5.0$\% frequency ($Head$), medium-category accuracy for those between 1.0\% and 5.0\% frequency ($Medium$), and tail-category accuracy for the remaining low-frequency behaviors ($Tail$). 

\subsection{Behavior prediction Metrics}
The formula for $Prec_w$ :
\begin{equation}\label{equ:Prec_w}
    Prec_w = \frac{\sum_{c \in C} (\text{TP}_c + \text{FP}_c) \cdot \text{Precision}_c}{\sum_{c \in C} (\text{TP}_c + \text{FP}_c)}
\end{equation}
The formula for $Rec_w$ :
\begin{equation}\label{equ:Rec_w}
    Rec_w = \frac{\sum_{c \in C} (\text{TP}_c + \text{FN}_c) \cdot \text{Recall}_c}{\sum_{c \in C} (\text{TP}_c + \text{FN}_c)}
\end{equation}
The formula for $Overall$ :
\begin{equation}\label{equ:Rec_m}
    Accuracy = \frac{1}{|C|} \sum_{c \in C} \frac{\text{TP}_c}{\text{TP}_c + \text{FN}_c}
\end{equation}
The formula for $Head$ :
\begin{equation}\label{equ:Rec_m}
    Accuary = \frac{1}{|C_h|} \sum_{c \in C_h} \frac{\text{TP}_c}{\text{TP}_c + \text{FN}_c}
\end{equation}
The formula for $Medium$ :
\begin{equation}\label{equ:Rec_m}
    Accuary = \frac{1}{|C_m|} \sum_{c \in C_m} \frac{\text{TP}_c}{\text{TP}_c + \text{FN}_c}
\end{equation}
The formula for $Tail$ :
\begin{equation}\label{equ:Rec_m}
    Accuary = \frac{1}{|C_t|} \sum_{c \in C_t} \frac{\text{TP}_c}{\text{TP}_c + \text{FN}_c}
\end{equation}

Where $|C|$ represents the total number of classes, $|C_h|$ represents the total number of classes belonging to the head category,
Where $|C_m|$ represents the total number of classes belonging to the medium category,
Where $|C_h|$ represents the total number of classes belonging to the tail category.
True Positives $\text({TP}_c)$ denotes the number of samples correctly classified as class $c$, False Positives $\text({FP}_c)$ represents the number of samples incorrectly classified as class $c$, and False Negatives $\text({FN}_c)$ stands for the number of samples incorrectly classified as other classes instead of class $c$. And $\text{Precision}_c$ and $\text{Recall}_c$ respectively refer to the precision and recall of class $c$.

\section{Details of Used Metrics for Generation}
\label{app:metric-gene}

The formula for $BLEU$:
\begin{equation}\label{equ:BLEU}
    \text{BLEU} = \text{BP} \cdot \exp\left(\sum_{n=1}^N w_n \log p_n\right)
\end{equation}
Where $\text{BP} = \min\left(1, e^{1 - r/c}\right)$ is the brevity penalty, $p_n$ is the modified $n$-gram precision, $r$ is the reference length, and $c$ is the candidate length.

The formula for $TVD$:
\begin{equation}\label{equ:TVD}
    \text{TVD}(P,Q) = \frac{1}{2} \sum_{i=1}^k |P(i) - Q(i)|
\end{equation}
Where $P$ and $Q$ are probability distributions over $k$ classes, $P(i)$ denotes the predicted probability of class $i$, $Q(i)$ denotes the ground truth probability.

The formula for $JSD$:
\begin{equation}\label{equ:JSD}
    \text{JSD}(P\|Q) = \sqrt{\frac{1}{2} D_{\text{KL}}(P\|M) + \frac{1}{2} D_{\text{KL}}(Q\|M)}
\end{equation}
Where $M = \frac{1}{2}(P+Q)$ is the midpoint distribution, and $D_{\text{KL}}$ denotes the Kullback-Leibler divergence:
\begin{equation}\label{equ:KLD}
    D_{\text{KL}}(P\|Q) = \sum_{i=1}^k P(i) \log\frac{P(i)}{Q(i)}
\end{equation}

\section{Details of Baselines}
\label{app:baselines}
\textbf{SASRec~\citep{kang2018self}.}
uses self-attention mechanisms to model user behavior sequences. It captures both short-term and long-term dependencies in sequential data, allowing it to focus on the most relevant items in the user's interaction history for recommendation.

\textbf{BehaveGPT~\citep{gong2025behavegpt}}
is a transformer-based model pre-trained specifically on user behavior data. Its novel pre-training paradigm enables it to learn complex behavior patterns and support various downstream tasks, including next behavior prediction, long-term generation, and cross-domain adaptation.

\textbf{PITuning~\citep{gong2024population}}
loads pre-trained Large Language Model (LLM) parameters to acquire textual knowledge and then designs an adaptive unlearning strategy to address the long-tail preference issue, achieving excellent performance in user behavior prediction.

\textbf{AlphaFuse~\citep{hu2025alphafuse}}
is a simple yet effective language-guided learning strategy that addresses long-tail intent modeling by learning ID embeddings within the null space of language embeddings.

\textbf{TALLRec~\citep{bao2023tallrec}}
is one of the earlier methods to integrate Large Language Models (LLMs) with the recommendation domain. It employs a two-stage tuning process—Alpaca Tuning and Rec-Tuning—to finetune LLMs for recommendations, enabling effective and efficient adaptation of LLMs with only a small number of tuning samples.

\textbf{A-LLMRec~\citep{kimlarge}}
bridges the knowledge between the language and recommendation domains by training an alignment network with a variety of tasks, targeting both warm and cold-start scenarios.

\textbf{CoLLM~\citep{zhang2025collm}}
captures collaboration information using external traditional models and maps it into the LLM's input embedding space as collaboration embeddings. This external integration allows effective modeling of collaboration without modifying the LLM, enabling flexible use of various collaboration modeling techniques.

\textbf{LLaRa~\citep{liao2024llara}}
introduces a hybrid prompting method that integrates both world knowledge and behavioral patterns into item representations. It conducts curriculum prompt tuning to achieve modality alignment.

For comparison, we also consider LLMs that are not fine-tuned on behavioral data, i.e., Deepseek-V3~\citep{deepseekai2025deepseekv3technicalreport}, which is a powerful Mixture-of-Experts (MoE) language model with 671B total parameters and 37B activated per token, offering performance comparable to GPT-4~\cite {openai2024gpt4ocard} at a lower cost.

\section{Implementation Details}
\label{app:Implementation}
The hardware used in this experiment consists of 8 NVIDIA A100 40G GPUs. We selected Qwen2.5-7B~\citep{qwen2025qwen25technicalreport} as the backbone for our experiments. Our experiments utilized the AdamW optimizer with a cosine annealing learning rate schedule, setting the warm-up proportion to 0.03. The maximum learning rate for cosine annealing was set to 5e-5, while both the minimum and initial warm-up learning rates were set to 1e-6. We conducted LoRA~\citep{hu2022lora} fine-tuning and parallel training acceleration. All experiments were performed with a maximum of 3 training epochs and a batch size of 96, selecting the best-performing model on the validation set for testing. Our experiments are typically completed within 8 hours.And for the experimental results, due to limited computing resources, we fixed the random seed to 42 and only ran it once.

\section{Details of Practical Applications}
\label{app:Practical Applications}

\subsection{Data Generation Process}
For the behavior dataset, we use our model and baselines to generate one day of user behavior data based on a history sequence of 100 behaviors (spanning over one day). From the generated output, we take the most recent 41 behaviors and use the first 40 to predict the final one.

\subsection{Downstream Task Experimental Settings}
To evaluate the utility of the generated synthetic data, we employed SASRec as the downstream behavior prediction model. To ensure a fair comparison across datasets of varying sizes (e.g., real data vs. real + synthetic data) and to address concerns regarding gradient steps, we adopted a ``train to convergence'' strategy. Instead of fixing the total number of gradient steps, we utilized \textbf{Early Stopping} with a patience of 5 epochs (monitoring validation loss). This approach ensures that all models, regardless of the training data volume, are trained to their maximum potential without overfitting or underfitting.
The specific hyperparameters used for the downstream SASRec model are consistent with standard settings and are detailed as follows:
\begin{itemize}
    \item \textbf{Model Architecture}: 
    \begin{itemize}
        \item Hidden Units: 50
        \item Number of Blocks (Layers): 2
        \item Number of Attention Heads: 1
        \item Dropout Rate: 0.1
        \item Max Sequence Length: 40
    \end{itemize}
    \item \textbf{Optimization}: 
    \begin{itemize}
        \item Optimizer: Adam
        \item Learning Rate: 0.001
        \item Batch Size: 16
        \item L2 Embedding Regularization (\texttt{l2\_emb}): 0.01
    \end{itemize}
    \item \textbf{Training Config}: 
    \begin{itemize}
        \item Maximum Epochs: 200
        \item Early Stopping Patience: 5 epochs
    \end{itemize}
\end{itemize}

\section{Ablation Study for Behavior Generation Task}
\label{app:Ablation-gene}
The ablation results on the generation task are shown in the following Table~\ref{tab:Ablation-gene}.
\begin{table*}[h]
\centering
\caption{Ablation Study for Behavior Generation Task}
\label{tab:Ablation-gene}
\makebox[\textwidth]{
        \scalebox{0.9}{
                \begin{tabular}{l *{3}{ccc}}
                \toprule
                \multirow{2}{*}{Method}  & 
                \multicolumn{3}{c}{Event} & 
                \multicolumn{3}{c}{Timestamp} & 
                \multicolumn{3}{c}{Location} \\
                \cmidrule(lr){2-4} \cmidrule(lr){5-7} \cmidrule(lr){8-10}
                 & $\text{Bleu} \uparrow$ & $\text{TVD} \downarrow$ & $\text{JSD} \downarrow$ & $\text{Bleu} \uparrow$ & $\text{TVD}\downarrow$ & $\text{JSD} \downarrow$ & $\text{Bleu} \uparrow$ & $\text{TVD} \downarrow$ & $\text{JSD} \downarrow$ \\
                \midrule
                Ours     & 0.354 & 0.140 & 0.020 & 0.541 & 0.147 & 0.020 & 0.711 & 0.065 & 0.005 \\
                w/o stage1         & 0.309 & 0.167 & 0.028 & 0.500& 0.162 & 0.024 & 0.640 & 0.093 & 0.007 \\
                w/o stage2  & 0.304 & 0.189 & 0.029 & 0.580 & 0.095 & 0.008 & 0.745 & 0.064 & 0.006\\ %
                w/o stage3  & 0.343 & 0.146 & 0.022 & 0.523 & 0.156 & 0.025 & 0.708 & 0.079 & 0.008  \\  
                \bottomrule
                \end{tabular}
        }
}

    \vspace{-5pt} 
\end{table*}

\section{Analysis of Error Sources in Behavior Prediction}
\label{app:error_source}
We additionally conducted error analysis experiments to better analyze the sources of error in the behavior prediction task. Specifically, for the three progressive subtasks in the second stage, we replaced the model-generated outputs with ground-truth values from their respective supervised training tasks. The experimental setup includes four groups, as shown below(Note that understanding task1, task2, and task3 in the table represent the User Key Behavior Identification, User Behavior Pattern, and User Feature Summarization Discovery tasks, respectively.). In the table, "pred" indicates that the corresponding feature uses the model’s own generated result (which may contain errors), while "label" denotes the use of the ground-truth value from the supervised training tasks.

\begin{table}[ht]  %
  \centering
  \caption{Experimental Setup for Error Analysis}  %
  \label{tab:error_analysis_setup}  %
  \setlength{\tabcolsep}{4pt}  %
  \resizebox{0.48\textwidth}{!}{  %
  \begin{tabular}{cccc}  %
    \toprule  %
    ID & \textit{understanding task1} & \textit{understanding task2} & \textit{understanding task3} \\
    \midrule  %
    1 & pred & pred & pred \\
    2 & label & pred & pred \\
    3 & label & label & pred \\
    4 & label & label & label \\
    \bottomrule  %
  \end{tabular}
  }
  \vspace{-5pt} 
\end{table}

Under these four experimental settings, we analyzed changes in the accuracy on the long-tail test set. The results are summarized in the table below. In the table, $r2w$ represents the percentage of data that changed from correct to incorrect predictions compared to the previous row's settings, while $w2r$ represents the opposite, and $\textit{Difference}$ indicates the net accuracy improvement (the difference between $w2r$ and $r2w$).

\begin{table}[ht]
  \centering
  \caption{Error Analysis Results on Long-Tail Test Set}  %
  \label{tab:error_analysis_results}  %
  \setlength{\tabcolsep}{6pt}  %
  \begin{tabular}{ccccc}  %
    \toprule
    ID & $Overall$ & $r2w$ & $w2r$ & $\textit{Difference}$ \\
    \midrule
    1 & 0.336 & - & - & - \\
    2 & 0.417 & 6.3\% & 14.4\% & 8.1\% \\
    3 & 0.432 & 2.7\% & 4.2\% & 1.5\% \\
    4 & 0.480 & 1.8\% & 6.6\% & 4.8\% \\
    \bottomrule
  \end{tabular}
\end{table}

The experimental results reveal that \textit{User Key Behavior Identification} and \textit{User Feature Summarization} have the greatest impact on errors. \textit{User Key Behavior Identification} serves as the starting point for behavioral analysis in stage 2, where even small initial errors can propagate and compound across subsequent subtasks. Meanwhile, the final \textit{User Feature Summarization} task, being directly linked to behavior prediction, significantly influences the final accuracy. The quality of the summarized features directly affects the precision of behavior predictions, hence its substantial impact. 

In the paper, we primarily focused on enhancing \textit{User Feature Summarization} through self-reflection optimization tasks. However, we acknowledge that insufficient attention was given to the \textit{User Key Behavior Identification} task, which also has a significant impact on errors. This insight offers a valuable direction for our future work.

\section{Joint Optimization} %
\textbf{Joint Optimization} – Consider using adaptive learning rate schedules to resolve convergence mismatch between prediction and generation tasks

We implemented an \textbf{adaptive learning rate schedule} by dynamically adjusting the task loss weights based on the ratio of current loss to initial loss. This effectively assigns a higher weight to the prediction task and a lower weight to the generation task, accelerating convergence of the former while slowing down the latter.
Below is a detailed description of the strategy:

\subsection{Dynamic Task Weighting Strategy}

Let:
\begin{itemize}
    \item $L^{(0)}_i$: the initial loss of task $i$
    \item $\hat{L}_i$: the current exponentially moving averaged (EMA) loss of task $i$
    \item $r_i = \frac{\hat{L}_i}{L^{(0)}_i}$: the loss ratio of task $i$
    \item $\bar{r} = \frac{1}{|\mathcal{V}|} \sum\limits_{i \in \mathcal{V}} r_i$: the average loss ratio across \textbf{valid tasks}
    \item $s_i = \frac{\bar{r}}{r_i}$: the \textbf{relative learning speed} of task $i$ (slower tasks will have larger values)
    \item $\alpha$: a tunable exponent to control the sensitivity of the weighting
\end{itemize}

The normalized task weight $w_i$ is computed as:  
\[
w_i = 
\begin{cases}
\frac{s_i^\alpha}{\sum\limits_{j \in \mathcal{V}} s_j^\alpha} \cdot |\mathcal{V}|, & \text{if } L^{(0)}_i > 0 \\
1, & \text{otherwise}
\end{cases}
\]  

Where:  
\[
\mathcal{V} = \left\{ i \mid L^{(0)}_i > 0 \right\}
\]  
is the set of valid tasks (i.e., those with positive initial loss values).

After applying this method, the step corresponding to the lowest total loss shifted from \textbf{2600 to 3200}, with corresponding unweighted prediction and generation losses improving slightly to \textbf{0.2674} and \textbf{0.271} (from \textbf{0.2690} and \textbf{0.2726}). These data show that this method does make the convergence speed of prediction and generation tasks more matched.

The table below shows performance comparisons, where "No optimization" refers to results without multi-task optimization (as in the paper)

\subsubsection{Prediction Task}  

\begin{table}[ht]
    \centering
    \caption{Performance of the Prediction Task Under Different Optimization Methods}
    \label{tab:prediction_optimization}
    \resizebox{0.48\textwidth}{!}{  %
    \setlength{\tabcolsep}{3pt} 
    \begin{tabular}{@{}lcccccc@{}}
        \toprule
        Optimization method    & $Prec_w$ & $Rec_w$ & $Overall$ & $Head$ & $Medium$ & $Tail$ \\
        \midrule
        No optimization        & 0.644    & 0.642   & 0.471     & 0.538  & 0.489    & 0.446  \\
        Adaptive learning rate & 0.638    & 0.648   & 0.479     & 0.528  & 0.497    & 0.452  \\
        \bottomrule
    \end{tabular}
    }
\end{table}

\subsubsection{Generation Task}

\begin{table*}[h]
\centering
\caption{Performance of the Generation Task Under Different Optimization Methods}
\label{tab:generation_optimization}
\makebox[\textwidth]{
        \scalebox{0.9}{
                \begin{tabular}{l *{3}{ccc}}
                \toprule
                \multirow{2}{*}{Method}  & 
                \multicolumn{3}{c}{Event} & 
                \multicolumn{3}{c}{Timestamp} & 
                \multicolumn{3}{c}{Location} \\
                \cmidrule(lr){2-4} \cmidrule(lr){5-7} \cmidrule(lr){8-10}
                 & $\text{Bleu} \uparrow$ & $\text{TVD} \downarrow$ & $\text{JSD} \downarrow$ & $\text{Bleu} \uparrow$ & $\text{TVD}\downarrow$ & $\text{JSD} \downarrow$ & $\text{Bleu} \uparrow$ & $\text{TVD} \downarrow$ & $\text{JSD} \downarrow$ \\
                \midrule
                None                 & 0.354 & 0.140 & 0.020 & 0.541 & 0.147 & 0.020 & 0.711 & 0.065 & 0.005 \\
                Adaptive learning rate & 0.363 & 0.141 & 0.020 & 0.553 & 0.146 & 0.019 & 0.708 & 0.079 & 0.008 \\
                \bottomrule
                \end{tabular}
        }
}
\end{table*}

As shown, while some metrics improved, results are not consistently better across all tasks. This suggests that \textbf{multi-task optimization requires more sophisticated strategies}, and we plan to explore further methods (e.g., separate optimizers or gradient balancing techniques) in future work.

\section{Efficiency Comparison}
\noindent \textbf{Efficiency Comparison} – Compare inference time and memory usage with baseline models

Regarding computational cost during inference:

\begin{itemize}
    \item \textbf{Hardware:} All inference tests were conducted on NVIDIA A100 (40GB).
    
    \item \textbf{Inference Time (on 20,000 samples from the Honor dataset):}
    \begin{itemize}
        \item \textit{$<3$ minutes:} SASRec, BehaveGPT, PITuning, AlphaFuse
        \item \textit{\textasciitilde $25$ minutes:} TALLRec, A-LLMRec, CoLLM, LLaRa
        \item \textit{\textasciitilde $40$ minutes:} BUA
        \item \textit{Not available:} DeepSeek (API-based)
    \end{itemize}
    
    \item \textbf{Memory Usage:}
    \begin{itemize}
        \item \textit{Low ($<2$GB):} SASRec (\textasciitilde 1GB), BehaveGPT, PITuning, AlphaFuse (\textasciitilde 2GB)
        \item \textit{High (\textasciitilde $30–32$GB):} TALLRec, A-LLMRec, CoLLM, LLaRa, BUA
    \end{itemize}
\end{itemize}

BUA’s inference efficiency is comparable to other LLM-based baselines, though higher than traditional methods—reflecting a broader trend in LLM-based approaches. We anticipate continued advances in LLM optimization that will help narrow this efficiency gap in the near future.

\section{Cross-Cultural Contexts Evaluation}
\noindent \textbf{Cross-Cultural Contexts Evaluation} – Test the model on datasets from different domains or cultural backgrounds

To address this, we incorporated a new dataset: the \textbf{Carat Top 1000 Users App Usage Dataset}, which collects app usage and battery data from volunteers across multiple countries, including the U.S., Japan, and the U.K., and notably excludes China. This provides a complementary perspective to the Honor and Tencent datasets used in our original submission.

We compared our method (BUA) with the best-performing baselines from each category on this dataset. The results are shown in Table \ref{tab:cross_cultural_results}.

\begin{table}[ht]
\centering
\caption{Performance on Carat Top 1000 Users App Usage Dataset}
\label{tab:cross_cultural_results}
    \resizebox{0.48\textwidth}{!}{  
        \setlength{\tabcolsep}{3pt} 
\begin{tabular}{lcccccc}
\toprule
Method         & $Prec_w$ & $Rec_w$ & $Overall$ & $Head$  & $Medium$ & $Tail$  \\
\midrule
BehaveGPT    & 0.299    & 0.318   & 0.210     & 0.303   & 0.219    & 0.052   \\
PITuning       & 0.352    & 0.356   & 0.357     & 0.362   & \underline{0.425}    & 0.152   \\
CoLLM          & \underline{0.400}    & \underline{0.365}   & \underline{0.367}     & \underline{0.377}   & 0.418    & \underline{0.233}   \\
Ours (BUA)     & \textbf{0.447}    & \textbf{0.418}   & \textbf{0.400}     & \textbf{0.409}   & \textbf{0.451}    & \textbf{0.267}   \\
\bottomrule
\end{tabular}
}
\end{table}

As shown, our method continues to achieve strong performance on a dataset with a markedly different demographic and geographical distribution, further validating its generalizability.

\section{Pretrained Base Model Replacement}
\noindent \textbf{Pretrained Base Model Replacement} – Evaluate the effect of replacing the current pre-trained base model

We have replaced BehaveGPT with SASRec as the pretrained behavior sequence encoder. The performance is shown in Table \ref{tab:pretrained_replacement}.

\begin{table}[ht]
    \centering
    \caption{Performance Comparison of Different Pretrained Base Models}
    \label{tab:pretrained_replacement}  %
        \resizebox{0.48\textwidth}{!}{  
        \setlength{\tabcolsep}{3pt} 
    \begin{tabular}{lcccccc}  %
        \toprule  %
        Pretrained Model  & $Prec_w$  & $Rec_w$  & $Overall$ & $Head$ & $Medium$ & $Tail$ \\
        \midrule  %
        SASRec            & 0.561     & 0.589    & 0.331     & 0.466  & 0.389    & 0.247  \\
        BehaveGPT       & 0.644     & 0.642    & 0.471     & 0.538  & 0.489    & 0.446  \\
        \bottomrule  %
    \end{tabular}
    }
\end{table}

While SASRec underperforms compared to BehaveGPT, our method still achieves notable gains over SASRec alone, demonstrating its effectiveness.

\section{Self-Reflection Details}
\noindent \textbf{Self-Reflection Details} – Provide a more detailed explanation of the self-reflection optimization method.

To clarify, the model identifies recurring shortcomings in initial profiles through prompt-guided reflection, focusing on issues such as:
\begin{enumerate}
    \item insufficient abstract summarization,
    \item inadequate detail association and reasoning,
    \item poor structural clarity,
    \item weak information hierarchy,
    \item inaccurate temporal pattern analysis, and
    \item lack of personalized expression.
\end{enumerate}

The corresponding correction criteria are designed as follows:
\begin{itemize}
    \item \textbf{For abstract summarization}: Elevate surface-level behaviors to infer deeper cognitive traits (e.g., deducing "information-driven lifestyle" from frequent news consumption).
    \item \textbf{For temporal analysis}: Calibrate behavior frequencies and highlight periodic patterns.
    \item \textbf{For structure}: Implement a three-layer hierarchy—from cognitive-level traits to habit interactions and specific behavioral anchors.
    \item \textbf{For personalization}: Emphasize distinctive, user-specific behavioral descriptors while avoiding vague generalities.
\end{itemize}

Importantly, this self-reflective process is not limited to output refinement. As described above, feedback from these reflections is also used to update model parameters via supervised fine-tuning, leading to further performance improvements.

\section{Data Granularity}
\noindent \textbf{Data Granularity} – Clarify what is meant by "high-level daily activities" and how they are represented in the data.

To clarify the granularity of “Behavior Type ID,” we define it at the level of high-level daily activities—neither raw sensor signals nor overly abstract categories. Below is a simplified example of a typical user’s day to illustrate the scope:

\begin{itemize}
    \item \textbf{Morning}: Alarm clock, check weather
    \item \textbf{Commute (to work)}: Subway, watch news, payment
    \item \textbf{Work hours}: Editing video, online meeting
    \item \textbf{Lunch break}: Ordering takeout, watching video
    \item \textbf{Commute (to home)}: Subway, watching video, payment
    \item \textbf{Evening}: Online shopping, gaming, watching video
\end{itemize}

Due to space limitations, this example condenses some activity details, but it reflects the typical granularity used across different scenarios.

\section{Human Evaluation}
\label{app:human_evaluation}
\noindent \textbf{Human Evaluation} – Human evaluation experiments on the quality and interpretability of model-generated profile features

We have conducted a human evaluation study to more systematically assess the interpretability of the generated portrait features.

Following the methodology of prior work [1], we randomly selected 120 test samples from the Honor dataset. For each sample, portrait features were generated by three sources: (1) human annotators, (2) our proposed model (BUA), and (3) the base model (Qwen2.5-7B without fine-tuning).
This resulted in a total of 360 portrait feature samples, which were evaluated by a separate group of human judges using consistent evaluation criteria.
The evaluation focused on two dimensions:
\begin{itemize} %
    \item \textbf{Rationality}: the degree to which the portrait features align with the user’s historical behavior
    \item \textbf{Interpretability}: the extent to which the portrait features help explain the user's predicted future behavior
\end{itemize}

Scores ranged from 0 to 3, with higher scores indicating better performance. To avoid bias, the order and source of the portrait features were anonymized and randomly shuffled for each evaluation instance. (Note: If multiple sources achieve the highest score for a sample, the credit is divided equally among them.)

\begin{table}[h]
    \centering
    \caption{Average Scores of Human Evaluation}
    \vspace{-3pt} 
    \label{tab:human_eval_avg}
    \resizebox{0.33\textwidth}{!}{
    \begin{tabular}{lcc}
        \toprule
        Type    & Rationality & Interpretability \\
        \midrule
        Human   & 2.51        & 2.55             \\
        BUA     & 2.46        & 2.39             \\
        No Tune & 1.90        & 1.83             \\
        \bottomrule
    \end{tabular}
    }
    \vspace{6pt}  
    
    \caption{Best Score Probability per Sample}
    \vspace{-3pt} 
    \label{tab:human_eval_best}
    \resizebox{0.48\textwidth}{!}{
    \begin{tabular}{lccc}
        \toprule
        Category                 & Human & BUA   & No Tune \\
        \midrule
        Best in Rationality      & 39.8\% & 37.9\% & 22.3\%   \\
        Best in Interpretability & 45.3\% & 33.3\% & 21.4\%   \\
        \bottomrule
    \end{tabular}
    }
    \vspace{-5pt} 
\end{table}

These results indicate that while human-written features still achieve the highest overall performance, our fine-tuned model (BUA) significantly outperforms the base model in both rationality and interpretability. Moreover, BUA’s performance approaches that of human-written features, demonstrating meaningful gains in interpretability.

\section{Questionnaire Setup Details}
\label{app:Questionnaire_setup}
\newcommand{\tasktitle}[1]{\subsubsection*{\textcolor{blue!70!black}{#1}}}
\newcommand{\criteriontitle}[1]{\subsection*{\textcolor{blue!80!black}{\large #1}}}

\tasktitle{Task Description:}
Your task is to evaluate the quality of user features based on two dimensions

\tasktitle{Dimension 1 (Rationality):}
Does the user feature accurately reflect the user's historical behavior sequence?

\tasktitle{Dimension 2 (Interpretability):}
Can the user feature help explain the predicted next behavior of the user? Does it provide a reasonable basis for why the predicted behavior might occur?

\vspace{5pt}
\hrulefill
\vspace{10pt}

\criteriontitle{Scoring Criteria for Dimension 1 (Rationality: 0–3 points)}
\begin{table}[ht]  %
    \centering
    \begin{tabular}{l p{0.40\textwidth}}  %
        \toprule
        \textbf{Score} & \textbf{Description} \\  %
        \midrule
        0 & \textbf{No Match}: The profile feature is not reflected at all in the user's behavior sequence. \\
        1 & \textbf{Weak Match}: The profile feature is only partially reflected in the behavior sequence. \\
        2 & \textbf{Basic Match}: The feature is generally reflected in the behavior sequence but is overly broad (e.g., ``user likes playing games''). \\
        3 & \textbf{Strong Match}: The feature is clearly and specifically reflected in the behavior sequence (e.g., ``user likes playing games on Friday nights after watching short videos''). \\
        \bottomrule
    \end{tabular}
\end{table}

\criteriontitle{Scoring Criteria for Dimension 2 (Interpretability: 0–3 points)}
\begin{table}[h]  %
    \centering
    \begin{tabular}{l p{0.38\textwidth}}
        \toprule
        \textbf{Score} & \textbf{Description} \\  %
        \midrule
        0 & \textbf{No Match}: The profile feature is completely unrelated to the predicted user behavior. \\
        1 & \textbf{Weak Match}: The feature can be loosely connected to the predicted behavior (e.g., ``user often engages in leisure activities'' → predicted behavior: ``playing games''). \\
        2 & \textbf{Basic Match}: The feature aligns with the predicted behavior but is too general (e.g., ``user likes playing games'' → predicted behavior: ``playing games''). \\
        3 & \textbf{Strong Match}: The feature directly and specifically supports the predicted behavior (e.g., ``user likes playing games on Friday nights after reading the news'' → predicted behavior: ``playing games''; it is Friday night and the user has just read the news). \\
        \bottomrule
    \end{tabular}
\end{table}

\begin{figure*}[t]  %
    \centering
    \includegraphics[width=0.8\textwidth]{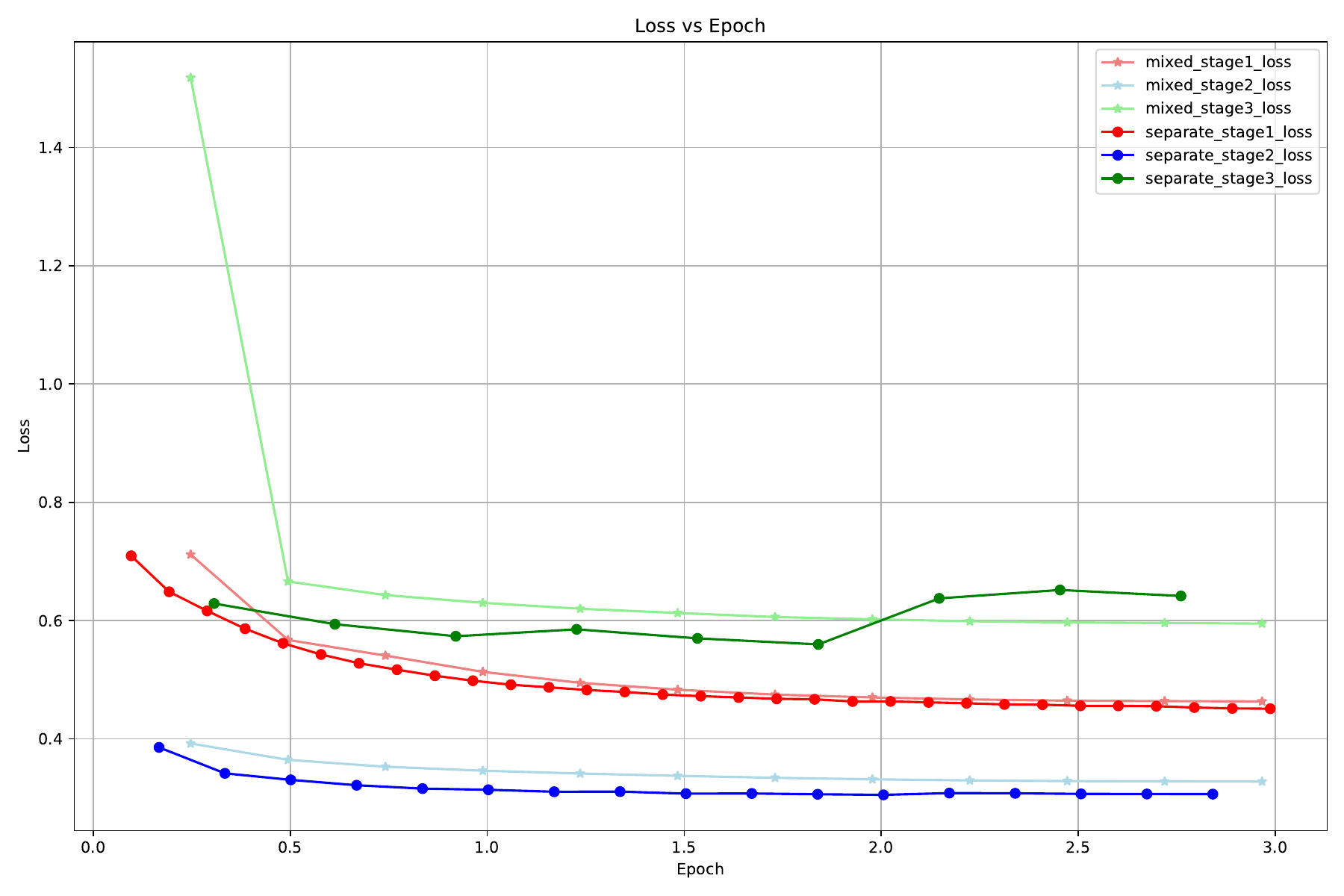} 
    \caption{Validation loss comparison between our Staged Curriculum (Separate) and Joint Training (Mixed) strategies. The solid lines represent our Staged Training, while the faded lines with stars represent Joint Training.}
    \label{fig:loss_curve}
\end{figure*}

\section{Empirical Analysis of Curriculum Learning Strategy}
\label{app:curriculum_analysis}
To empirically validate the necessity and effectiveness of our proposed three-stage curriculum design, we conducted a comprehensive ablation study comparing the convergence dynamics of our approach against a standard multi-task learning baseline. Specifically, we contrasted our proposed \textbf{Staged Training} strategy, where the model is optimized sequentially through Sequence-Level Understanding (Stage 1), User-Level Feature Modeling (Stage 2), and Self-Reflective Refinement (Stage 3), against a \textbf{Joint Training} baseline. In the Joint setting, the model is trained simultaneously on all tasks across the three stages from scratch, disregarding the hierarchical dependencies inherent in behavioral understanding. This comparison aims to verify whether the structured, easy-to-hard progression provides tangible optimization benefits over simple joint optimization.

The validation loss curves for each stage's specific tasks under both settings are presented in Figure~\ref{fig:loss_curve}. The comparative analysis reveals three critical insights regarding the training dynamics:
\begin{itemize}
    \item \textbf{Comparable Performance on Baselines (Stage 1):} For the most fundamental task, Sequence-Level Understanding, the loss curves for both Separate (Red solid line) and Mixed (Red faded line) settings are relatively close. This indicates that simple semantic alignment is less sensitive to the training strategy and can be adequately learned via joint optimization.
    \item \textbf{Superiority in User-Level Modeling (Stage 2):} A significant divergence appears in the more difficult Stage 2 tasks. The Separate Training (Blue solid line) achieves a consistently lower minimum validation loss compared to the Mixed setting (Blue faded line). This confirms that a solid foundation in Stage 1 is essential for mastering complex user features, as the model benefits from pre-aligned semantic representations.
    \item \textbf{Cold-Start Challenge in Self-Reflection (Stage 3):} Notably, the Mixed\_Stage3\_Loss (Green faded line) starts at an extremely high value ($>1.5$), indicating that the model struggles to perform self-reflective refinement without a pre-established user profile context. In contrast, the Staged approach (Green solid line) allows the model to tackle Stage 3 with initialized understanding, resulting in a smoother optimization landscape and lower final loss.
\end{itemize}

These empirical results strongly corroborate the theoretical foundation of our curriculum design, rooted in cognitive scaffolding and curriculum learning (Bengio et al., 2009). The convergence patterns demonstrate that while joint training is sufficient for aligning basic semantic representations, it struggles with higher-order reasoning tasks without established prerequisites. By enforcing a structured, easy-to-hard learning progression, our staged approach ensures that the model acquires a robust understanding of fundamental behavioral semantics before tackling complex user profiling and self-reflection, achieving superior performance.

\section{Use of LLMs}
We used LLMs to assist in writing the paper, such as identifying typos and correcting grammatical errors, as well as polishing some paragraphs.

\section{Synthetic Data for Downstream Prediction Task}
\label{app:Synthetic_data}
\begin{figure}
    \centering
    \includegraphics[width=1\linewidth]{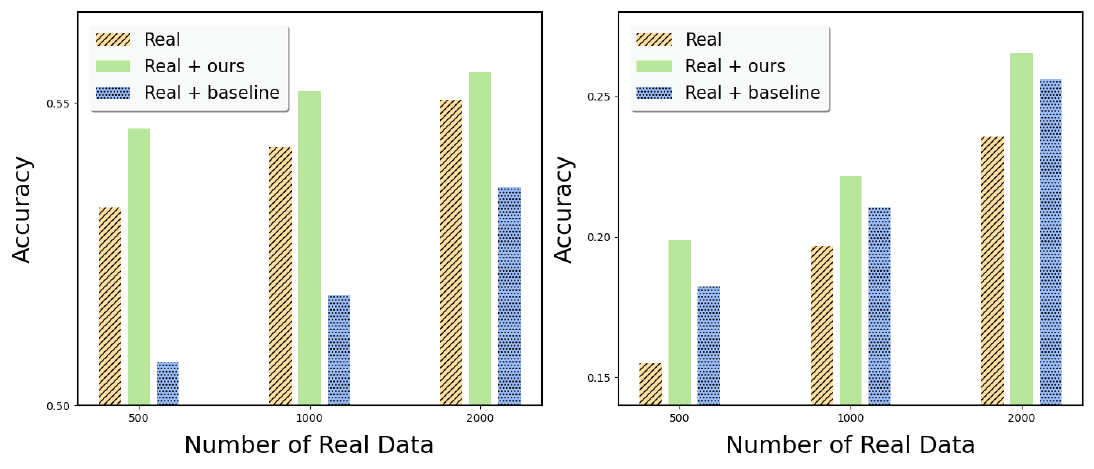}
    \caption{hybrid scenario}
    \label{fig:enter-label}
\end{figure}
To further assess the usability of the generated data, following ~\citep{yuan2023learning}, we evaluated our model in a hybrid setting that augments real data with synthetic data. Using the standard SASRec model for next-behavior prediction (see Appendix~\ref{app:Practical Applications}). As shown in Figure~\ref{fig:enter-label}, our generated data consistently outperforms the strongest baseline, D2A, in both overall and average accuracy, significantly boosting model performance. This confirms our model’s ability to generate high-fidelity behavior sequences that capture underlying user patterns.

\end{document}